# Predictive Allostatic Organization in Recurrent and Spiking Agents Under Partial Observability

**Frederick Hayes III**
Independent Researcher
San Antonio, Texas, USA
ORCID: https://orcid.org/0009-0009-5492-0105
fhayes3@berkeley.edu

## Abstract

Adaptive behavior under partial observability depends on internal organization that carries information beyond the current observation. Drawing on Barrett and Miller's account of categorization as predictive, compressive, functionally organized, and allostatically constrained, we test whether recurrent and spiking agents develop internal states with corresponding computational properties. Agents operate in an energy-constrained foraging task requiring resource acquisition, threat avoidance, contact-dependent consumption, and regulation of an internal energy variable. In a frozen benchmark, learned agents outperform random and heuristic baselines; the trace-augmented recurrent policy is strongest overall, while spiking variants show stress-specific differences. Early internal dynamics predict later full-safe-efficient success above permutation baseline, reaching a maximum ROC-AUC of 0.802. Reduced PCA subspaces retain behaviorally relevant information. Feature-family controls show that predictive signal is distributed across trace, policy-head, internal-dynamics, observation, and allostatic variables, and low-energy state remains strongly decodable after explicit energy-related features are removed. Evaluation-time perturbations to temporal state, sensory information, operating conditions, and allostatic mechanisms alter behavior and/or internal prediction. Seed-balanced event probes show weaker but measurable information about future contact, successful consumption, and threat events, alongside strong low-energy decoding. We interpret this pattern as a computational analogue of predictive allostatic organization: distributed control regimes that are predictive, energy-sensitive, action-relevant, and partly causally involved, without claiming biological validation or discrete symbolic categories.

## 1. Introduction

Adaptive behavior under uncertainty requires more than selecting an action from the current observation. In partially observable environments, the meaning of an observation depends on recent history, current goals, internal resources, and likely future states. A contact signal may indicate an opportunity only if the agent has reached a resource. A threat cue may matter differently depending on internal energy and distance from reward. A consume action can be efficient, wasteful, or dangerous depending on the state in which it is selected. In such settings, endpoint performance alone does not tell the whole story. The internal organization that makes adaptive behavior possible becomes part of what must be explained.

This paper studies that internal organization in a controlled artificial-agent setting. We evaluate recurrent and spiking agents in an energy-constrained foraging task with partial observability, resource acquisition, threat avoidance, contact sensing, and an internal energy variable. The goal is not to produce a new large-scale reinforcement-learning benchmark or to claim state-of-the-art performance. Instead, we ask a mechanistic question: can learned agents form internal dynamics that predict future-relevant outcomes, compress task variation, discriminate functional contexts, encode energy-sensitive state, and contribute causally to adaptive behavior?

The theoretical motivation comes from Barrett and Miller's account of categorization as a fundamental operating principle of brain function. They argue that categorization is not merely the final stage of perception, where a completed sensory representation is assigned to a stored label. Rather, categorization is a process by which organisms treat objects, actions, or events as equivalent for a purpose. In their account, predictive feedback organizes incoming signals from the outset, dimensionality reduction compresses high-dimensional variation into lower-dimensional functional structure, and allostasis links this organization to energetic regulation and action. (Barrett & Miller, 2026)

We do not attempt to model or validate that biological theory. Instead, we use it to derive operational criteria for artificial agents. If an artificial system exhibits a computational analogue of predictive allostatic categorization, then its internal dynamics should do more than mirror the current observation. They should predict later outcomes, preserve behaviorally relevant information under compression, discriminate action-relevant contexts, carry energy-sensitive information, and be causally involved in control. These criteria define the empirical program of the paper.

This distinction is important because artificial-agent research often evaluates policies primarily by external performance: reward, completion rate, failure rate, or sample efficiency. Those metrics are necessary, but they do not reveal whether successful behavior depends on organized internal control regimes or brittle action mappings. Conversely, bio-inspired artificial intelligence often invokes mechanisms such as spiking dynamics, allostasis, oscillation, or predictive processing without directly testing whether those mechanisms matter. The present study is designed to sit between these extremes. It uses a deliberately small task environment so that internal state, perturbations, and outcome structure can be measured directly.

The task is a compact foraging world in which agents acquire resources, avoid threats, manage internal energy, and decide when to consume. Agents receive partial and noisy observations, including resource and threat cues, energy state, time, and contact information. We compare random and heuristic baselines against learned recurrent and spiking controllers. The recurrent model provides a strong memory-based baseline under partial observability. The spiking models expose mechanistic variables such as membrane persistence, spike activity, recurrent current, operating temperature, gain, threshold modulation, and allostatic gating. (Bellec et al., 2018; Neftci et al., 2019; Eshraghian et al., 2023)

The study proceeds through five linked experiments: a frozen behavioral benchmark, internal-state prediction and compression analyses, feature-family and allostatic controls, evaluation-time perturbations, and a seed-balanced future-event analysis. Together, these experiments test whether recurrent and spiking agents form internal dynamics that are predictive, compressive, energy-sensitive, functionally discriminative, and mechanistically involved in control.

Across these experiments, learned agents form distributed predictive control regimes: internal states predict later success, encode energy-sensitive structure, and support action-relevant discrimination. The strongest interpretation is bounded: the agents form predictive, energy-sensitive internal control states, not discrete symbolic categories or biological models.

This paper makes four contributions. First, it translates Barrett and Miller's predictive allostatic account into operational criteria for artificial agents: prediction, compression, functional discrimination, allostatic sensitivity, and causal involvement. Second, it introduces a controlled energy-constrained partial-observability benchmark for recurrent and spiking agents. Third, it provides an internal-state analysis pipeline that separates internal dynamics from reward, outcome labels, action-use metrics, and other behavior-derived variables. Fourth, it combines held-out behavioral evaluation, feature-family controls, allostatic controls, causal perturbations, and future-event probes to test whether internal dynamics are not only predictive, but mechanistically relevant to adaptive control.

The remainder of the paper introduces the conceptual bridge, defines the task and methods, presents the results, and discusses limitations and future directions.

# 2. Conceptual Framework and Related Work

## 2.1 From category labels to functional equivalence

This study draws on a long shift in categorization research away from the view that categories are fixed symbolic containers assigned after perception is complete. Classical accounts often treated categorization as a process in which perceptual features are extracted, compared with stored concepts, and then assigned to a class. Later work on prototypes, exemplars, graded membership, and goal-derived categories complicated that picture by showing that categories can be context-sensitive, purpose-dependent, and organized around what an organism or agent needs to do. (Bruner et al., 1956; Rosch, 1975; Medin & Schaffer, 1978; Nosofsky, 1986; Barsalou, 1983, 1991)

That shift matters for artificial agents because resource approach, contact opportunity, threat context, low energy, and imminent consume events are functional situations. What matters is whether internal dynamics organize sensory histories into states useful for action.

Embodied and situated approaches make a related point: cognition unfolds through interaction between an agent and its environment rather than through detached internal description alone. (Agre & Chapman, 1987; Brooks, 1991; Varela et al., 1991; Clark, 1997; Gallagher, 2023) The present work treats internal state as meaningful insofar as it participates in control: preserving temporal context, anticipating future-relevant conditions, regulating internal energy, and guiding action under uncertainty.

This framing is especially relevant under partial observability, where the same cue can require different behavior depending on recent history and internal condition. The bridge to the present experiments is therefore functional: category-like structure is operationalized as organization of internal control states, not symbolic labeling.

## 2.2 Prediction, allostasis, and Barrett–Miller's synthesis

Predictive-processing and active-inference traditions add a second strand to the argument. They frame perception and action as prospective processes: an adaptive system does not merely wait for sensory input and respond, but maintains internal dynamics that anticipate future sensory, bodily, or action-relevant conditions. (Friston, 2010; Clark, 2013; Hohwy, 2013; Seth, 2013) Allostasis extends this predictive logic to energetic regulation. A reactive system restores variables after they deviate; an allostatic system prepares for expected needs before they fully arise. (Sterling, 2012; Sterling & Laughlin, 2015; Barrett & Simmons, 2015; Schulkin & Sterling, 2019)

Barrett and Miller provide the central synthesis for this paper: categorization is treated as predictive, compressive, functionally organized, and allostatically constrained, rather than as late-stage label assignment.

The present paper does not attempt to validate that biological account. Rather, it uses the account to derive operational criteria for an artificial system. If an artificial agent exhibits a computational analogue of predictive allostatic categorization, then its internal dynamics should satisfy five conditions:

1. **Prediction:** internal dynamics should anticipate later success or future task events.
2. **Compression:** reduced internal subspaces should preserve behaviorally relevant information.
3. **Functional discrimination:** internal state should distinguish action-relevant contexts such as contact opportunity, resource approach, threat context, and low energy.
4. **Allostatic sensitivity:** energy-relevant information should be represented beyond raw observation alone.
5. **Causal involvement:** perturbing candidate mechanisms should alter behavior and/or internal predictive structure.

These criteria define both the empirical program and the boundary of the claim: the question is whether agents form distributed internal control regimes with analogous properties, not biological categories or symbolic concepts.

**Table 1. From Barrett–Miller theory to operational agent tests**

Table 1 summarizes the theoretical bridge used throughout the paper by converting Barrett and Miller's central commitments into measurable criteria for artificial agents.

| Barrett–Miller concept | Operational analogue in this study | Empirical test |
|---|---|---|
| Categorization as functional equivalence | Different sensory/task situations become internally distinguishable by action-relevant role rather than by symbolic label | Functional-state probes for contact opportunity, resource approach, threat context, low energy, and future-event classes |
| Predictive feedback / prospective organization | Internal state contains information about later success or near-future events before those outcomes occur | Early-window probes for later full-safe-efficient success; seed-balanced future-event probes |
| Compression / dimensionality reduction | Task-relevant information remains available in reduced internal subspaces | PCA and compression probes over instrumented internal dynamics |
| Allostasis / energetic constraint | Internal dynamics carry energy-sensitive information beyond raw observation alone | Low-energy decoding; energy-control probes excluding explicit energy/allostatic variables |
| Causal involvement | Perturbing internal mechanisms changes behavior and/or internal predictive structure | Evaluation-time perturbations of trace, state persistence, membrane persistence, temperature, contact, recurrent current, and allostatic modulation |
| Guardrail | The system forms distributed control regimes, not biological categories or clean symbolic concepts | Discussion of weak/mixed geometric category separation and weak-to-moderate exact event prediction |

This mapping defines the scope of the paper's claims: evidence requires more than behavioral success alone, so the remaining experiments test prediction, compression, functional discrimination, allostatic sensitivity, and perturbation sensitivity.

Figure 1. Study overview and operational bridge
Energy-constrained foraging task -> recurrent/spiking agents -> operational tests of predictive allostatic organization
A. Task: energy-constrained partial observability
Goal: consume resources, avoid threats, preserve energy
agent
Internal energy constrains action
energy[t]
Partial observations: resource cue + threat cue + energy + time + contact
B. Agents and instrumented internal state
Trace-augmented observation base obs + decaying context
Baselines random policy; greedy heuristic
Trace-GRU policy hidden state h[t] policy logits
Spiking policies membrane V[t] · spikes S[t] recurrent current · temperature allostatic gain/threshold
Instrumented state trace · core dynamics · policy head · allostatic variables
Evaluation targets full-safe-efficient success · failures · future events
C. Theory-to-evidence bridge
Barrett-Miller inspiration prediction · compression functional equivalence · allostasis
Prediction
Compression
Functional discrimination
Allostatic sensitivity
Causal involvement
Prospective event information
Evidence streams behavior · probes/PCA · controls perturbations · event probes
Claim tested: distributed predictive control regimes under partial observability
Abbreviations: FSE = full-safe-efficient success; trace = decaying task-context features; allostatic = engineered energy-sensitive control variable.

**Figure 1. Study overview and operational bridge.** The task is an energy-constrained partially observable foraging environment with resources, threats, contact-dependent consume actions, and internal energy. Recurrent and spiking agents are evaluated through behavioral benchmarks, internal-state probes, feature-family controls, causal perturbations, and seed-balanced future-event analyses. Barrett and Miller's predictive allostatic account is translated into operational criteria for artificial agents: prediction, compression, functional discrimination, allostatic sensitivity, causal involvement, and prospective event information.

## 2.3 Recurrent and spiking agents as temporal control systems

The computational side of the study begins from a practical observation: partial observability makes memory necessary. When the current observation does not fully specify the task state, a policy must preserve information across time. Recurrent neural networks provide a standard way to do this: hidden state can integrate recent observations, carry task context, and approximate belief over unobserved variables. (Kaelbling et al., 1998; Hausknecht & Stone, 2015; Ni et al., 2021)

Spiking neural networks provide a different temporal substrate. In a spiking model, computation is shaped not only by static activation values but also by membrane state, spike events, thresholds, recurrent currents, and temporal dynamics. These variables create additional ways for history and internal condition to influence action. (Bellec et al., 2018; Neftci et al., 2019; Eshraghian et al., 2023; Zhou et al., 2024)

The present study treats recurrent and spiking agents as complementary tools. The trace-augmented recurrent model provides a strong non-spiking memory baseline, while the spiking models expose perturbable temporal and energy-sensitive variables such as membrane persistence, recurrent current, spike summaries, allostatic gain, threshold modulation, operating temperature, and oscillatory gating.

Bio-inspired mechanisms become explanatory only if they contribute to predictive internal structure, behavior, or perturbation sensitivity; adding spiking dynamics, oscillations, or energy-like variables is not sufficient by itself.

### 2.4 Internal-state analysis and the operational gap

External performance alone cannot establish whether an agent has formed meaningful internal organization. A policy may succeed because it has learned a robust internal control regime, but it may also succeed through brittle action mappings, shortcut cues, or selection artifacts. Conversely, a policy may fail while still forming internally separable states that predict its own collapse. To study predictive allostatic organization, we therefore measure both behavior and internal dynamics.

This requires three methodological commitments. First, the analysis must record dynamic state rather than static model attributes. For recurrent policies, this includes hidden-state summaries and trace-augmented inputs. For spiking policies, it includes membrane summaries, spike summaries, recurrent/spiking state variables, policy-head quantities, and allostatic modulation summaries. Second, the feature matrices must avoid behavioral-label circularity: reward, return, resource counts, success labels, consume precision, action-use metrics, and other behavior-derived variables are excluded from internal probes. Third, decodability must be complemented by perturbation. If an internal state predicts success, that does not by itself show causal involvement; perturbing candidate mechanisms tests whether they matter for behavior and/or internal predictive structure.

The gap addressed by this paper is methodological as much as conceptual: what is less common is a controlled workflow that connects functional-equivalence theory, embodied action, predictive allostasis, temporal artificial agents, internal-state probing, and perturbation.

The experiments that follow implement this workflow by testing behavioral competence, internal prediction and compression, feature-family and allostatic controls, perturbation sensitivity, and prospective event information. Together, they evaluate whether recurrent and spiking agents under partial observability form predictive allostatic control regimes in the computational sense defined above.

## 3. Task, Models, and Evaluation Protocol

### 3.1 Environment and outcome taxonomy

We evaluate agents in a compact energy-constrained foraging environment designed to make internal state necessary but interpretable. Each episode takes place in a 7 × 7 gridworld containing an agent, three resources, and three threats. The agent must navigate the grid, acquire resources, avoid threats, manage an internal energy variable, and decide when to execute a consume action. Episodes terminate when all resources are consumed, energy is depleted, or the 40-step horizon is reached.

The base observation has seven components:

o[t] = [resource_dx[t], resource_dy[t], threat_dx[t], threat_dy[t], energy[t], step_fraction[t], contact[t]]

The first two components encode the noisy relative direction to the nearest unconsumed resource. The next two encode the noisy relative direction to the nearest threat. The fifth component is normalized energy, the sixth is normalized episode progress, and the seventh is a binary contact sensor indicating whether the agent is standing on an unconsumed resource. The agent does not receive a full map of the environment.

The action set is:

A = {up, down, left, right, stay, consume}

Movement actions update the agent's position subject to boundary constraints. The consume action succeeds only when the contact condition is active. Successful consume actions increase reward, increase the number of consumed resources, and restore energy. Failed consume actions are penalized. Threat contact imposes a penalty and reduces energy. The internal energy variable is an engineered resource constraint: it makes internal condition relevant to control, but it is not intended as a biological metabolism model.

To test robustness, we evaluate agents across a frozen suite of partial-observability variants. These variants alter the reliability or availability of resource cues, threat cues, contact sensing, or combinations of these signals. They include base observation, delayed contact, resource-cue dropout, resource-and-threat dropout, no-contact, resource-dropout/no-contact, severe partial observability, and combined-stress variants.

The primary behavioral outcome is full-safe-efficient success. This outcome requires full task completion, safety, and efficient consume behavior. Full completion requires the episode-level full-completion indicator to be active. Safety requires fewer than three threat hits; episodes with three or more threat hits are marked unsafe. Efficiency requires at least one resource to be consumed, consume precision of at least 0.50, and no more than 3.0 consume attempts per successful consume. Secondary outcomes include full completion, partial success, no-resource failure, mean reward, consume precision, threat hits, and final energy.

**Table 2. Task and evaluation overview**

Table 2 summarizes the task and evaluation protocol used for all main results. The environment is deliberately compact, but it combines partial observability, resource acquisition, threat avoidance, contact-dependent action, and an internal energy constraint. These design choices make the task small enough to instrument while still requiring temporal integration.

| Component | Final protocol value |
| --- | --- |
| Environment | Energy-constrained partially observable foraging gridworld |
| Grid size | 7 × 7 |

| Component | Final protocol value |
|---|---|
| Resources / threats | 3 resources, 3 threats |
| Episode horizon | 40 steps |
| Actions | up, down, left, right, stay, consume |
| Base observation | resource cue, threat cue, energy, step fraction, contact sensor |
| Primary outcome | full-safe-efficient success |
| Safety threshold | threat_hits < 3 |
| Efficiency threshold | consume precision >= 0.50 and actions per success <= 3.0 |
| Stress variants | cue dropout, delayed contact, no contact, severe partial observability, combined stress |
| Frozen behavioral rows | 1.8 million |

The table gives the minimal experimental specification for the main manuscript; full variant definitions, outcome-label logic, model registry details, and seed construction are provided in the appendices.

## 3.2 Model families and trace augmentation

We compare non-learning baselines, recurrent learned agents, and spiking learned agents. The random policy provides a lower-bound baseline. The greedy heuristic provides a hand-coded solvability baseline using resource cues, contact sensing, and simple threat avoidance. The primary recurrent learned model is a trace-augmented GRU policy, which maintains hidden state over time and provides a strong non-spiking memory baseline under partial observability.

The spiking models use recurrent state variables and energy-sensitive modulation. They expose internal quantities such as membrane state, spike summaries, recurrent current, gain, threshold modulation, gate variables, operating temperature, and policy-head output. Several spiking operating points are evaluated, including behavior-cloned and reinforcement-learning-trained variants with different train/evaluation temperatures.

All learned policies receive trace-augmented observations in the primary conditions. The augmented observation is:

o_aug[t] = concat(o[t], z[t])

where o[t] is the seven-dimensional base observation and z[t] is the trace state. The trace state is updated from the current observation and its previous value:

z[t] = f_trace(o[t], z[t-1])

The update function includes exponential decay and cue-derived updates, allowing recent task-relevant information to persist across steps without giving the agent a full map of the environment. In later analyses, trace features are treated as their own feature family so that their contribution can be separated from recurrent/spiking core dynamics, policy-head quantities, observation features, and explicit allostatic variables.

The comparison between recurrent and spiking models is not designed to show that spiking models are globally superior. The recurrent model provides the strongest general memory baseline, while the spiking models provide a mechanistic testbed for temporal and energy-sensitive variables that can be instrumented and perturbed.

## 3.3 Frozen evaluation protocol and internal-state instrumentation

All headline results come from a frozen evaluation sequence established after pilot development. Pilot experiments were used to stabilize the environment, identify candidate mechanisms, and freeze the final protocol; they are not used as final quantitative evidence.

The frozen behavioral benchmark evaluates each model across 20 training seeds, 20 final test base seeds, 50 evaluation episodes per test base seed, and 10 frozen environment variants. This gives 1,000 held-out evaluation episodes per trained model/variant instance and 1.8 million behavioral evaluation rows across the benchmark.

Internal-state analyses use structured subsets and derived samples from the frozen benchmark. These analyses preserve the same model registry, seed discipline, and variant definitions while reducing computational cost for instrumentation and probing.

Dynamic internal-state instrumentation records the state used by the deployed policy during evaluation. For recurrent models, recorded quantities include GRU hidden-state summaries, trace-augmented inputs, policy logits, action probabilities, entropy, and confidence summaries. For spiking models, recorded quantities include membrane-state summaries, spike summaries, recurrent/spiking current summaries, threshold and gain modulation, gate-related variables, allostatic state summaries, policy logits, and action probabilities.

The analyses distinguish several feature families:

- trace-only

- core dynamics
- core dynamics without explicit energy terms
- explicit allostatic variables
- policy-head features
- internal dynamics
- internal dynamics without explicit energy terms
- internal-all features
- observation-only
- observation without energy

Representation analyses exclude reward, total return, outcome labels, resource counts, consume precision, action-use summaries, and other behavior-derived metrics from internal feature matrices. These variables are used as labels or evaluation targets, not as internal explanatory features. This separation is necessary to avoid behavior-label circularity.

## 3.4 Probes, perturbations, and event analyses

We use supervised probes to test whether internal features predict task-relevant targets. The primary target is later full-safe-efficient success. Additional targets include no-resource failure, full completion, contact opportunity, resource approach, threat context, low-energy state, future contact, future successful consume, and future threat hit.

The primary probe metric is ROC-AUC, supplemented by balanced accuracy and average precision where appropriate. Probes are evaluated using grouped validation by training seed, so that train/test splits do not mix examples from the same trained seed group. Permutation-label baselines are used for key targets.

Compression is evaluated using PCA over internal-dynamics features. PCA is not treated as a complete account of internal geometry. It is used as a conservative compression test: whether lower-dimensional projections preserve behaviorally relevant information.

To test causal involvement, we apply evaluation-time perturbations to trained policies. These perturbations alter deployed mechanisms without retraining the models. They include zeroing trace features, zeroing contact-trace features, resetting recurrent/spiking state, combining trace removal with state reset, neutralizing energy observation, zeroing contact sensor input, disabling allostatic modulation, disabling oscillatory gating, disabling recurrent current, removing membrane persistence, and evaluating spiking policies at a changed operating temperature.

Finally, we test prospective event information using seed-balanced event samples drawn from evaluation trajectories and organized by training seed, model, perturbation, and environment variant. Future-event targets use a five-step look-ahead horizon for upcoming contact, successful consumption, and threat hits, while energy-sensitive state is monitored using an internal-energy threshold of 8.0.

## 4. Results

The experiments evaluate whether recurrent and spiking agents satisfy the operational criteria defined in Section 2. We report the main findings here and move full tables, secondary metrics, and fold-level probe results to the appendices.

### 4.1 Learned agents outperform baselines under a frozen behavioral benchmark

The frozen behavioral benchmark produced a clear separation between random behavior, heuristic control, and learned recurrent/spiking policies. The random policy served as the lower-bound baseline and failed to complete the task under the final protocol. The greedy heuristic performed substantially better than random, confirming that the environment contains exploitable structure, but remained below the learned agents.

The strongest overall model was the trace-augmented recurrent policy, Trace-GRU. Across hard variants, it achieved the highest mean full-safe-efficient success rate and the best aggregate robustness score. Several spiking variants were close behind, including RL T0.10, BC T0.10, and Spiking Default. The behavioral result therefore does not show that spiking agents globally outperform recurrent agents. Instead, it shows that recurrent and spiking agents occupy a meaningful learned-control tier above random and heuristic baselines, with trace-augmented recurrence strongest overall and spiking policies showing stress-specific structure.

The benchmark also revealed boundary conditions. The base-observation variant was nearly solved, with the best model reaching full-safe-efficient success of 0.981 and full completion of 0.998, but performance varied under delayed contact, cue dropout, severe partial observability, and contact removal. Combined-stress no-contact was effectively unsolved, showing that the agents learn robust but limited control regimes rather than a general solution to all partial-observability conditions.

**Figure 2. Frozen behavioral benchmark**

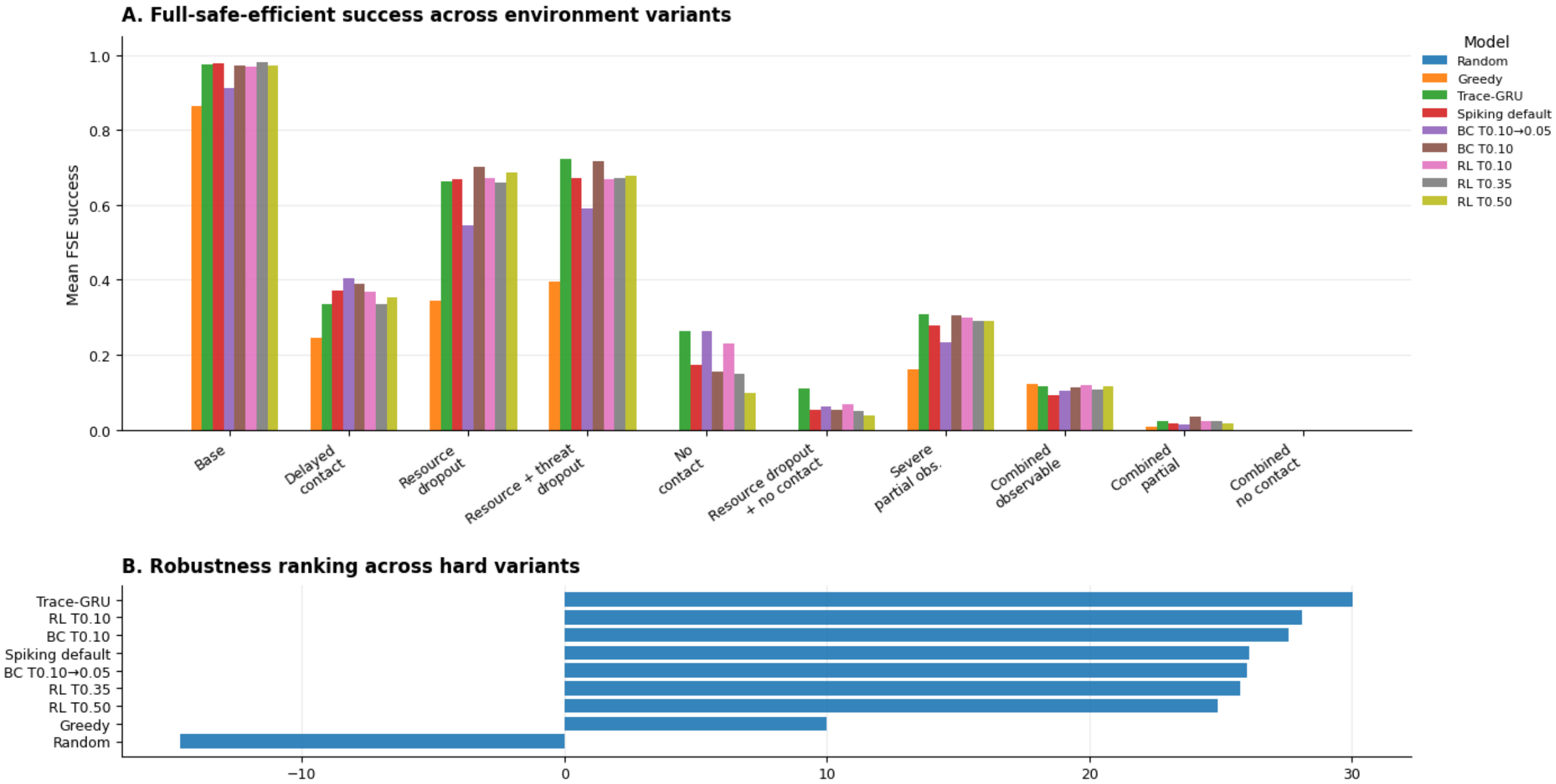

**Figure 2. Frozen behavioral benchmark.** Mean full-safe-efficient success across model families and environment variants shows a learned-control tier above random and heuristic baselines. The trace-augmented recurrent policy is strongest overall across hard variants, while spiking variants show stress-specific structure. Contact removal and combined-stress variants expose boundary conditions in which available observations and learned mechanisms are insufficient for robust control.

Full behavioral rankings and best-by-variant tables are reported in Appendix B.

## 4.2 Early internal dynamics predict later success and preserve information under compression

We next asked whether behavioral regimes were reflected in internal dynamics. The final instrumented representation subset contained 15,000 episode-level rows and 1,530 columns, including 1,431 internal feature columns and 42 observation feature columns. This confirmed that the analysis captured dynamic policy state rather than only static model metadata.

Early-window internal dynamics predicted later full-safe-efficient success above permutation baseline for all analyzed learned models. The strongest result was observed in Spiking Default, with ROC-AUC = 0.802 compared with a permutation mean of 0.498. Trace-GRU also showed above-chance prediction, with ROC-AUC = 0.715 compared with a permutation mean of 0.503. Across models, deltas over permutation ranged from +0.212 to +0.304.

Thus, early internal state contains prospective information about later behavioral success rather than merely reflecting the current observation.

**Figure 3. Internal prediction and compression**

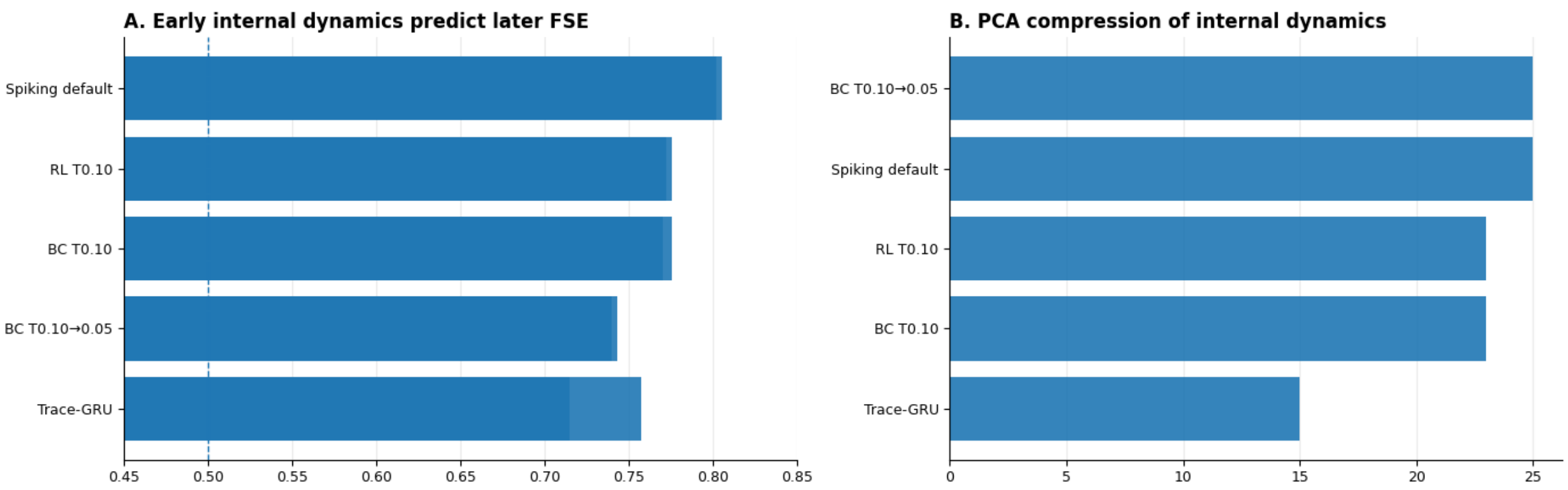

**Figure 3. Internal prediction and compression.** Early internal dynamics predict later full-safe-efficient success above chance across learned models. The dashed line marks chance-level ROC-AUC. PCA compression shows the number of principal components needed to explain 80% of internal-dynamics variance. The recurrent model has more compact measured structure than the spiking variants, while spiking models show stronger early outcome decodability in some cases.

The internal prediction results did not simply reproduce the behavioral ranking. Trace-GRU was strongest overall behaviorally, but Spiking Default produced the strongest early internal-dynamics FSE prediction. This dissociation is useful: behavioral competence and linear decodability are related but not identical. Behavioral success depends on action selection and environmental interaction, while internal decodability measures how separable later outcomes are from the measured state.

Compression analyses showed that behaviorally relevant information is preserved in reduced internal subspaces. The trace-GRU model had the most compact PCA structure: its first two components explained 38.7% of variance, and 15 dimensions reached 80% cumulative variance. Spiking models required more dimensions, typically around 23–25 dimensions for 80% variance. Reduced PCA subspaces remained predictive of final success: for example, Trace-GRU reached ROC-AUC = 0.706 using 16 principal components, Spiking Default reached 0.762, and BC T0.10 → eval T0.05 reached 0.813.

These results support the prediction and compression criteria.

## 4.3 Predictive information is distributed across trace, dynamics, policy output, and allostatic state

To test whether internal prediction was reducible to a single feature source, we decomposed the representation into feature families. Trace-only features were the strongest early predictors of full-safe-efficient success, with mean ROC-AUC = 0.841. Policy-head-only features were also strongly predictive, with mean ROC-AUC = 0.791. Internal-all features reached 0.768, internal-dynamics features reached 0.760, and observation-only features were also predictive at approximately 0.757. Core dynamics alone were weaker but still above chance, reaching approximately 0.660; after explicit energy terms were removed, core dynamics remained predictive at approximately 0.628.

The main result is that predictive information is distributed. Trace features are especially strong, policy-head quantities are action-proximal and informative, and recurrent/spiking core dynamics retain signal even under stricter controls. This prevents a simple interpretation in which the internal result is only a policy-head artifact or only a raw-observation effect.

Energy-control probes further strengthened the allostatic result. Low-energy state remained strongly decodable after explicit energy-related variables were removed. Explicit allostatic variables produced mean low-energy ROC-AUC = 0.934, as expected. More importantly, internal dynamics without explicit energy terms reached approximately 0.927, and core dynamics without explicit energy terms reached approximately 0.886. This indicates that energy-relevant structure is distributed through internal dynamics rather than confined to a single explicit energy channel.

Functional-state probes showed that internal dynamics discriminate action-relevant contexts. Contact opportunity was strongly decodable, with internal-dynamics ROC-AUC = 0.984. Low-energy state reached ROC-AUC = 0.949. Threat context and resource approach reached 0.845 and 0.830, respectively. Future contact within five steps was weaker but above chance at 0.629.

At the same time, geometric functional-equivalence tests remained weak or mixed. Internal states support supervised functional discrimination, but they do not appear as clean, compact category clusters under simple geometric metrics. This distinction is central: the learned agents form distributed control regimes, not symbolic category islands.

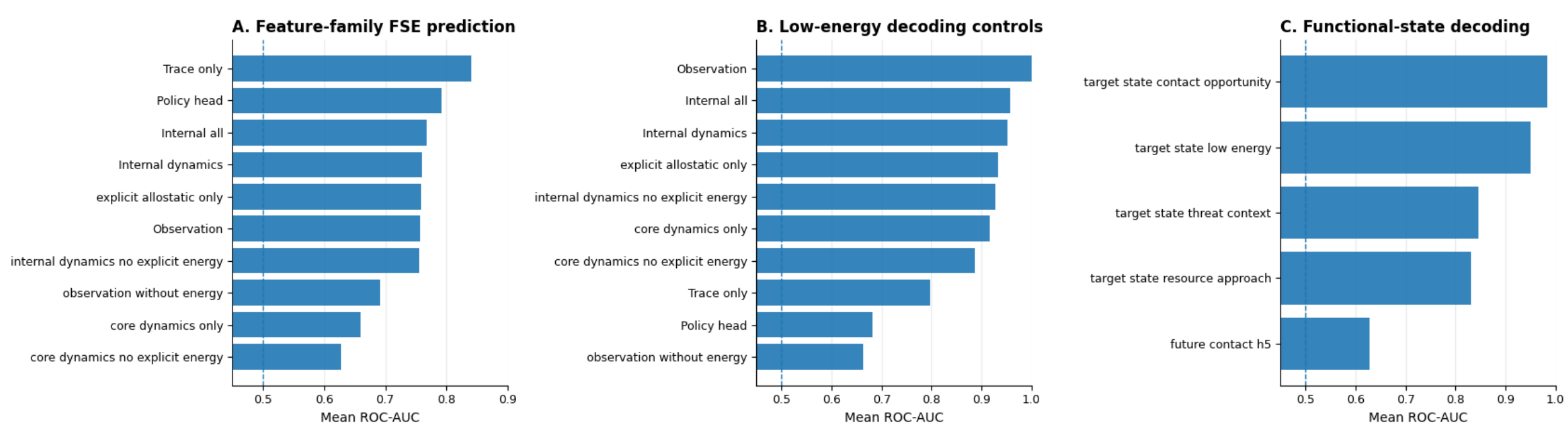

**Figure 4. Feature-family, allostatic, and functional controls.** Predictive information is distributed across trace, policy-head, internal-dynamics, observation, and allostatic variables. Low-energy state remains strongly decodable after explicit energy and allostatic terms are removed, indicating that energy-relevant structure is distributed through internal dynamics rather than confined to a single raw input channel. Internal dynamics also discriminate functional contexts including contact opportunity, low energy, threat context, resource approach, and near-future contact.

Full feature-family tables, energy-control results, functional probes, held-out-variant tests, and geometry analyses are reported in Appendix B.

## 4.4 Evaluation-time perturbations identify mechanisms that affect behavior and internal prediction

To test causal involvement, we applied evaluation-time perturbations to trained policies. These perturbations targeted trace information, contact information, recurrent/spiking state persistence, spiking membrane dynamics, recurrent current, allostatic modulation, oscillatory gating, and spiking operating temperature.

Several perturbations produced large behavioral effects. Removing spiking membrane persistence caused the strongest aggregate degradation, reducing mean full-safe-efficient success by -0.447, increasing no-resource failure by +0.430, and reducing reward by -22.36. Evaluating spiking policies at temperature 1.00 also produced strong degradation, with mean FSE delta -0.408 and reward delta -20.60. Zeroing contact information reduced mean FSE by -0.373. Trace perturbations also mattered: zeroing all trace features reduced mean FSE by -0.313, while the combined trace-removal/state-reset perturbation reduced mean FSE by -0.240.

The perturbation results were not uniform. Allostatic modulation and oscillatory gating produced smaller, model-specific behavioral effects. Energy-neutral observation affected internal prediction more than behavior. This pattern suggests that multiple mechanisms contribute to control, but they do so at different levels: membrane persistence, operating temperature, and contact information produce the largest behavioral collapses, while trace and state persistence provide broader temporal support.

We then asked whether behavioral degradation was accompanied by weaker internal predictive structure. Across the perturbation suite, 13 of 55 model–perturbation summaries showed both a full-safe-efficient success drop and an early internal-dynamics AUC drop. The strongest combined effects included membrane persistence removal in Spiking Default, temperature changes in several spiking models, contact removal in Trace-GRU, trace-plus-state lesions, recurrent-current removal, and allostatic-modulation removal in selected spiking models.

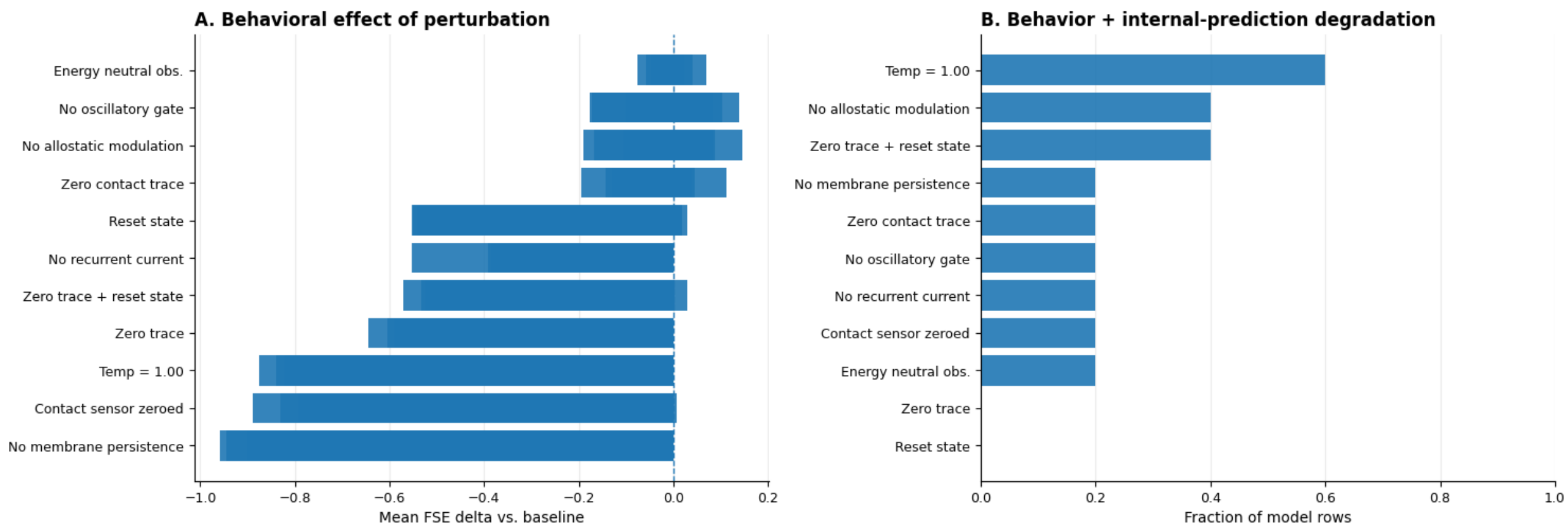

**Figure 5. Evaluation-time causal perturbations.** Perturbations to membrane persistence, spiking operating temperature, contact information, trace, persistent state, recurrent current, and allostatic modulation affect behavior and/or internal prediction. Panel A shows mean full-safe-efficient success change relative to baseline. Panel B shows the fraction of model–perturbation rows in which both behavioral success and early internal-dynamics prediction degrade. These results support causal involvement for several deployed mechanisms while also showing that behavioral degradation and decodability are not identical.

Thus, perturbation results support causal involvement for specific mechanism–model combinations while reinforcing that decodability and competence are not identical.

Supplemental perturbation summaries are reported in Appendix B; full model–perturbation rows are retained in the output bundle.

## 4.5 Seed-balanced event probes show weak-to-moderate prospective event information

Finally, we tested whether internal dynamics predict specific upcoming task events. Event-level samples were drawn from saved evaluation trajectories and balanced by training seed × model × perturbation × environment variant. Each group was capped at 125 rows. The final seed-balanced event matrix contained 825,000 rows and 312 columns, built from 19,800 event-sample files. It preserved 6,600 seed/model/perturbation/variant groups.

All event targets were evaluable. Future contact within five steps had 407,094 positives and 417,906 negatives. Future successful consume had 320,552 positives and 504,448 negatives. Future threat hit had 211,345 positives and 613,655 negatives. Current low energy had 86,052 positives and 738,948 negatives.

Baseline internal-dynamics probes showed a clear hierarchy. Current low-energy state was strongly decodable, with ROC-AUC = 0.914. Exact future events were weaker: future contact reached ROC-AUC = 0.602, future successful consume reached 0.648, and future threat hit reached 0.577.

These results clarify the level at which the internal representation operates. The agents do not precisely forecast local events. Instead, their internal dynamics carry modest prospective information about upcoming contact, consume, and threat events, with much stronger representation of energy state and broader success regimes. This supports the interpretation that internal state functions more like a predictive control-regime representation than an exact event predictor.

Event prediction was also perturbation-sensitive. Using a >0.02 ROC-AUC reduction threshold relative to baseline, 82% of perturbation rows reduced future successful-consume prediction, and 78% reduced future threat-hit prediction. This suggests that the future-event signal depends on the same broad family of temporal and allostatic mechanisms implicated in behavior and episode-level prediction.

**Figure 6. Seed-balanced future-event analysis**

**Figure 6. Seed-balanced future-event analysis.** Internal dynamics weakly-to-moderately predict future contact, successful consume, and threat-hit events, while low-energy state is strongly decodable. Perturbation sensitivity shows that event-level predictive information depends on temporal and allostatic mechanisms implicated elsewhere in the study. The event results support prospective control-regime information, but exact future-event prediction remains weaker than broader regime-level prediction.

Event-target coverage, event probe summaries, and perturbation-delta tables are reported in Appendix B.

## 4.6 Results summary

Across the five result blocks, learned agents outperform baselines; early internal dynamics predict later success and preserve information under compression; predictive information is distributed across trace, dynamics, policy-head, observation, and allostatic variables; perturbations affect behavior and internal prediction; and event probes show weak-to-moderate prospective event information with strong low-energy decoding.

Together, these findings support a bounded interpretation: recurrent and spiking agents form distributed predictive control regimes under partial observability, not clean symbolic categories or exact future-event predictors.

# 5. Discussion

## 5.1 Distributed predictive control regimes

The experiments support a bounded but meaningful interpretation: under energy constraint and partial observability, learned recurrent and spiking agents develop distributed internal control regimes. These regimes are predictive, compressive, energy-sensitive, action-relevant, and partly perturbation-sensitive.

The central result is not symbolic categorization: functional geometry is weak or mixed, and exact event prediction is weaker than broader regime-level prediction. The internal organization is better understood as graded control structure that organizes sensory histories by implications for action, energy, and future outcome.

This interpretation fits the operational bridge introduced in Section 2, where Barrett and Miller define categorization in terms of functional equivalence for a purpose rather than symbolic labeling. (Barrett & Miller, 2026) In the present system, functional equivalence appears as predictive control structure.

## 5.2 Regime-level prediction is stronger than exact event prediction

A consistent pattern is that internal dynamics predict broad outcomes more strongly than exact near-future events. Early internal dynamics predict later full-safe-efficient success above permutation baseline, and low-energy state is strongly decodable; future contact, successful consume, and threat-hit prediction are weaker.

The observed pattern sits between detailed event forecasting and purely reactive action mapping. Internal dynamics carry future-relevant task information, but that information is coarser than exact event timing.

The best interpretation is that the agents represent control regimes rather than precise futures: approach, contact opportunity, low-energy urgency, or drift toward failure can guide action without specifying exactly when an event will occur.

The event results therefore support the broader predictive-allostatic claim while limiting it: the system is predictive at the level of future-relevant control and energetic consequence more than exact event timing.

## 5.3 Trace, energy, and state persistence

Three mechanisms are especially important for interpreting the results: trace-linked context, energy-sensitive state, and temporal persistence. Trace features strongly predict later success and trace perturbations degrade behavior, especially when combined with state resets. The trace vector is not a full environment memory, but its limited decaying history is sufficient to support predictive context under partial observability.

Low-energy state is also strongly decodable, including after explicit energy and allostatic variables are removed. Although the energy variable is engineered, the learned policies organize internal state around an energy-like constraint.

Temporal persistence is most visible in perturbation results: membrane persistence, operating temperature, state reset, and recurrent-current disruption affect behavior and/or internal prediction, showing that the temporal substrate matters for deployed control.

Thus, the recurrent and spiking models play complementary roles: the trace-GRU is the strongest behavioral baseline, while the spiking models expose perturbable temporal and allostatic mechanisms.

## 5.4 Decodability is not adaptive quality

The perturbation results clarify what decodability can and cannot show. Probes reveal separable information in measured features, but not whether that information is adaptively used; some perturbations reduce both behavior and predictive structure, while others reduce behavior without reducing decodability.

A damaged policy may become more predictable by collapsing into a stereotyped failure regime; in that case, internal state is predictive but not adaptive.

For this reason, the paper treats decodability as evidence of internal structure, not as a direct measure of intelligence or control quality, and interprets it jointly with behavior, controls, and perturbations.

## 5.5 Relation to Barrett and Miller

Barrett and Miller's account motivates the paper, but the present study does not validate their biological theory. Their account concerns brain organization, cortical feedback, compression, interoception, and allostasis; this system is a small artificial task with engineered energy, trace features, and recurrent/spiking policies.

The value of the comparison is operational. Barrett and Miller suggest that categorization can be predictive, compressive, action-relevant, and allostatically grounded; this paper asks what those properties look like in an artificial agent, where they can be tested as prediction, compression, functional discrimination, allostatic sensitivity, causal involvement, and prospective event information.

Under those criteria, the system shows meaningful but bounded evidence. That is why the phrase computational analogue is important: the system is not a biological model, but an engineered testbed for asking whether predictive allostatic organization can be operationalized in artificial agents.

**Table 3. Evidence matrix for predictive allostatic organization.** The table summarizes how each operational criterion derived from Barrett and Miller's account is tested in the artificial-agent system, the level of support observed, and the main interpretive guardrail.

| Operational criterion | Primary evidence | Status | Interpretation | Main guardrail |
|---|---|---|---|---|
| Behavioral competence under partial observability | Learned recurrent and spiking agents outperform random and heuristic baselines under the frozen behavioral protocol; Trace-GRU is strongest overall, with spiking variants showing stress-specific structure. | Supported | The task suite produces meaningful learned-control regimes rather than random or purely heuristic behavior. | This is not a claim of spiking superiority; the recurrent model is strongest overall. |
| Prediction | Early internal dynamics predict later full-safe-efficient success above permutation baseline across analyzed learned models. | Supported | Internal state contains prospective information about later behavioral outcome. | Decodability indicates separable internal structure, not |

| Operational criterion | Primary evidence | Status | Interpretation | Main guardrail |
|---|---|---|---|---|
| | | | | automatically adaptive use. |
| Compression | PCA and reduced-subspace probes preserve behaviorally relevant information about later success. | Supported | Predictive information is not dependent on the full measured feature space. | PCA is a conservative compression probe, not a full account of internal geometry. |
| Feature-family specificity | Trace-only, policy-head, internal-dynamics, observation, and allostatic feature families each carry predictive signal. | Supported | Predictive information is distributed across trace, dynamics, action-proximal, and energy-sensitive variables. | Feature families are not independent mechanisms by themselves. |
| Allostatic sensitivity | Low-energy state remains strongly decodable, including after explicit energy/allostatic variables are removed. | Supported | Energy-relevant structure is distributed through internal dynamics rather than confined to the raw energy channel. | The energy variable and allostatic modulation are engineered task/architecture components, not biological metabolism. |
| Functional discrimination | Internal dynamics discriminate contact opportunity, low energy, resource approach, threat context, and near-future contact. | Supported with caveat | Internal state distinguishes action-relevant contexts. | Supervised functional probes show decodability; geometric category separation remains weak/mixed. |
| Causal involvement | Evaluation-time perturbations to trace, state persistence, membrane persistence, temperature, contact, recurrent current, and allostatic modulation affect behavior and/or internal prediction. | Partially supported | Several deployed mechanisms are behaviorally and representationally consequential. | Evaluation-time perturbations test deployed dependence, not retrained necessity. |
| Prospective event information | Seed-balanced event probes show weak-to-moderate prediction of future contact, successful | Weak-to-moderate support | Internal state carries limited prospective event information, especially around consume/contact structure. | The agents represent broad control regimes more strongly |

| Operational criterion | Primary evidence | Status | Interpretation | Main guardrail |
|---|---|---|---|---|
| | consume, and threat hit; low-energy state is strongly decoded. | | | than exact future-event timing. |
| Overall claim | Across behavior, probes, controls, perturbations, and event analysis, agents form distributed predictive control regimes. | Supported as computational analogue | The system operationalizes predictive allostatic organization in a controlled artificial-agent setting. | This is not biological validation and not evidence of clean symbolic categories. |

### 5.6 Summary

The contribution is an operational workflow for testing a theoretically motivated claim about internal organization. By linking frozen behavioral evaluation, dynamic internal-state instrumentation, probes, feature-family controls, perturbations, and event analysis, the paper shows how to evaluate whether agents form predictive, action-relevant, energy-sensitive control regimes.

These regimes are category-like only in the functional sense: they organize variation into states useful for action and regulation, but they are not discrete symbolic categories, exact future-event predictors, or biological models.

## 6. Limitations and Future Work

### 6.1 Limitations

The results support a bounded claim: recurrent and spiking agents in a controlled energy-constrained task can develop internal dynamics with properties analogous to predictive allostatic organization. They do not show biological categorization, discrete symbolic categories, or exact predictive world models.

The first limitation is environmental scope. The 7 × 7 foraging gridworld makes instrumentation, perturbation, and interpretation tractable, but limits generality; the results should be read as evidence from a controlled testbed, not proof that the same organization will scale to richer embodied, continuous-control, or open-ended tasks.

The second limitation is architectural scaffolding. Energy, trace, and allostatic modulation are engineered components, not blank-slate discoveries. The models still had to learn how to use them, and perturbations show that they matter, but the paper should not claim that allostatic or trace-like organization emerged without scaffolding.

The third limitation concerns model scope. The comparison is sufficient for the reported mechanistic tests, but it does not survey the full space of recurrent, spiking, transformer-style, neuromorphic, or hybrid architectures. The spiking models are valuable because they expose perturbable mechanisms, not because they dominate the trace-GRU baseline.

The fourth limitation is causal scope. Evaluation-time lesions test whether deployed mechanisms matter after learning, but they are not equivalent to retraining ablations and may create out-of-distribution policy states. The results therefore support claims about deployed dependence, not necessity for learning from scratch.

The fifth limitation concerns representation analysis. Supervised probes, PCA, and functional-state labels reveal measurable structure, but they do not fully explain how information is used. Because distributed control regimes may not appear as compact clusters, the paper interprets decodability jointly with behavior and perturbation rather than as a direct measure of adaptive quality.

The sixth limitation is event prediction. Future contact, successful consume, and threat-hit targets are evaluable, but prediction remains weak-to-moderate, suggesting that learned dynamics represent broad control regimes more strongly than exact short-horizon event forecasts.

Finally, the paper does not report hardware energy measurements. The task energy variable is not compute energy, and the spiking policies are not deployed on neuromorphic hardware.

## 6.2 Future work

Future work should first test whether the same internal organization appears in richer environments, including larger grids, moving threats, delayed rewards, multiple resource types, continuous control, and learned affordances.

A second direction is to compare engineered trace against learned memory, including no-trace agents, learned trace, adaptive decay, attention-like recurrence, eligibility traces, and other memory mechanisms.

A third direction is to replace evaluation-time perturbations with matched retraining ablations, distinguishing mechanisms necessary for deployed control from mechanisms necessary for learning.

A fourth direction is richer dynamical analysis: future work should study internal states as trajectories using transition models, nonlinear manifold methods, causal mediation, probe transfer, and dynamical-systems tools.

A fifth direction is to refine event prediction by testing multiple horizons, event timing, action-conditioned event likelihood, and event-balanced sampling around rare threat and consume transitions.

Hardware-aware evaluation should also measure spike count, event sparsity, memory movement, latency, numerical precision, and energy estimates on neuromorphic or FPGA-oriented platforms.

Finally, the theoretical bridge should be expanded across categorization theory, embodied cognition, predictive processing, interoception, recurrent control, spiking computation, and representation analysis while preserving the central guardrail: artificial agents can be evaluated for predictive allostatic organization without claiming biological equivalence.

## 6.3 Conclusion

This paper began from a theoretical question inspired by Barrett and Miller's account of predictive allostatic categorization: can an artificial agent under uncertainty and energetic constraint form internal dynamics that organize task variation into predictive, action-relevant, energy-sensitive structure?

Across five linked experiments, the answer is partially but meaningfully yes: learned agents outperform baselines; internal dynamics predict later success, preserve compressed information, encode low-energy state, discriminate functional contexts, and carry weak-to-moderate future-event information; and perturbations show that several mechanisms affect behavior and/or internal predictive structure.

The final interpretation remains bounded: the agents form distributed predictive control regimes, not clean symbolic categories, exact world models, or biological theories.

The contribution is a controlled workflow for studying artificial systems not only by what they do, but by how their internal dynamics organize future-relevant action under uncertainty.

# Appendix A. Task, Models, and Protocol Details

Appendix A provides compact audit details for the task, model families, outcome taxonomy, and frozen evaluation protocol used in the main manuscript. The full project output bundle contains lower-level logs and complete derived tables; this appendix includes only the information needed to interpret and audit the main text.

## A.1 Task configuration

| Component | Specification |
|---|---|
| Environment | Energy-constrained partially observable foraging gridworld |
| Grid size | 7 x 7 |
| Resources / threats | 3 resources and 3 threats per episode |
| Episode horizon | 40 steps |
| Actions | up, down, left, right, stay, consume |
| Base observation | resource cue, threat cue, energy, step fraction, contact sensor |
| Primary outcome | full-safe-efficient success |
| Safety threshold | safe if threat_hits < 3 |

| Component | Specification |
| --- | --- |
| Efficiency threshold | consume precision >= 0.50 and actions per success <= 3.0 |
| Behavioral evaluation rows | 1.8 million |

## A.2 Environment variants

| Variant | Modification | Purpose |
| --- | --- | --- |
| Base observation | Base observation stream. | Learnability reference condition. |
| Delayed contact | Contact signal delayed by two steps. | Tests temporal/contact memory under delayed contact signal. |
| Resource cue dropout | Resource cue dropped on 50% of eligible steps. | Tests resource-directed behavior under resource-cue uncertainty. |
| Resource + threat dropout | Resource and threat cues each dropped on 50% of eligible steps. | Tests control when both opportunity and hazard cues are unreliable. |
| No contact sensor | Contact sensor removed from observation. | Tests consume decisions when direct contact sensing is removed. |
| Resource dropout + no contact | Resource cue dropout combined with no contact sensor. | Tests joint resource uncertainty and missing contact information. |
| Severe partial observability | Severe cue degradation / partial observability stress. | Broad stress condition with stronger observation degradation. |
| Combined stress observable | Combined stress condition with contact observable. | Combined stress with contact still observable. |
| Combined stress partial | Combined stress condition under partial observability. | Combined stress under partial observability. |
| Combined stress no contact | Combined stress condition with no contact sensor. | Boundary condition combining stress with no contact signal. |

## A.3 Outcome taxonomy

| Quantity / label | Operational definition |
|---|---|
| Full completion | Episode-level full-completion indicator is active. |
| Partial completion | At least one resource is consumed, or partial-success indicator is active, without full completion. |
| No-resource failure | No resource is consumed. |
| Safe | threat_hits < 3. |
| Unsafe | threat_hits >= 3. |
| Efficient | At least one resource is consumed, consume precision >= 0.50, and actions per success <= 3.0. |
| Inefficient | At least one resource is consumed but one or more efficiency criteria are not met. |
| Full-safe-efficient success | Full completion AND safe AND efficient. |

## A.4 Model families

| Model family | Type | Role |
|---|---|---|
| Random | Non-learning baseline | Lower-bound chance behavior. |
| Greedy heuristic | Hand-coded baseline | Solvability reference using cue-following, contact sensing, and simple threat avoidance. |
| Trace-GRU | Recurrent learned policy | Strong memory baseline under partial observability. |
| Spiking default | Spiking learned policy | Primary spiking policy with membrane/spike state and allostatic variables. |
| BC / RL temperature variants | Spiking operating points | Tests sensitivity to training/evaluation temperature and behavior-cloned vs. RL-trained operating regimes. |

## A.5 Frozen evaluation protocol

| Protocol quantity | Value |
|---|---|
| Training seeds | 20 |
| Final test base seeds | 20 |
| Evaluation episodes per test base seed | 50 |
| Episodes per model/variant instance | 1,000 |
| Frozen environment variants | 10 |
| Final behavioral evaluation rows | 1,800,000 |

# Appendix B. Supplemental Results Tables

Appendix B provides compact supplemental tables supporting the main empirical claims. Full fold-level outputs, complete model-by-perturbation rows, and raw event-level artifacts are retained in the project output bundle rather than reproduced in the PDF.

## Table B1. Overall robustness ranking across hard variants.

| Model | Robustness score |
|---|---:|
| Trace-GRU | 30.021 |
| RL T0.10 | 28.116 |
| BC T0.10 | 27.608 |
| Spiking default | 26.09 |
| BC T0.10 -> eval T0.05 | 25.991 |
| RL T0.35 | 25.731 |
| RL T0.50 | 24.893 |
| Greedy heuristic | 9.998 |
| Random | -14.622 |

## Table B2. Best model by environment variant.

| Variant | Best model | Mean FSE | Full completion | No-resource failure | Mean reward |
|---|---|---|---|---|---|
| Base observation | RL T0.35 | 0.981 | 0.998 | 0 | 32.205 |
| Combined stress no contact | Random | 0 | 0 | 0.871 | -19.042 |
| Combined stress observable | Greedy heuristic | 0.122 | 0.122 | 0.285 | -10.799 |
| Combined stress partial | BC T0.10 | 0.034 | 0.036 | 0.519 | -14.874 |
| Delayed contact | BC T0.10 -> eval T0.05 | 0.405 | 0.562 | 0.044 | 12.513 |
| No contact sensor | Trace-GRU | 0.264 | 0.305 | 0.275 | -0.019 |
| Resource + threat dropout | Trace-GRU | 0.722 | 0.817 | 0.01 | 22.681 |
| Resource cue dropout | BC T0.10 | 0.702 | 0.781 | 0.021 | 20.109 |
| Resource dropout + no contact | Trace-GRU | 0.109 | 0.177 | 0.313 | -6.1 |
| Severe partial observability | Trace-GRU | 0.309 | 0.359 | 0.062 | 4.252 |

## Table B3. Early internal-dynamics prediction of later FSE success.

| Model | ROC-AUC | Permutation mean | Delta over permutation |
|---|---|---|---|
| Spiking default | 0.802 | 0.498 | +0.304 |
| RL T0.10 | 0.775 | 0.500 | +0.276 |

| BC T0.10 | 0.770 | 0.497 | +0.273 |
|---|---|---|---|
| BC T0.10 -> eval T0.05 | 0.740 | 0.498 | +0.241 |
| Trace-GRU | 0.715 | 0.503 | +0.212 |

## Table B4. PCA summary of early internal-dynamics features.

| Model | PC1 variance | PC2 variance | Dims for 50% variance | Dims for 80% variance |
|---|---:|---:|---:|---:|
| Trace-GRU | 0.222 | 0.166 | 5 | 15 |
| BC T0.10 | 0.134 | 0.11 | 7 | 23 |
| RL T0.10 | 0.135 | 0.104 | 8 | 23 |
| Spiking default | 0.131 | 0.116 | 8 | 25 |
| BC T0.10 -> eval T0.05 | 0.136 | 0.095 | 8 | 25 |

## Table B5. Feature-family prediction of later FSE success.

| Feature family | Mean FSE ROC-AUC |
|---|---:|
| Trace only | 0.841 |
| Policy head only | 0.791 |
| Internal all | 0.768 |
| Internal dynamics | 0.76 |
| Explicit allostatic variables | 0.758 |
| Observation | 0.757 |
| internal dynamics no explicit energy | 0.755 |
| Observation without energy | 0.691 |
| Core dynamics | 0.66 |
| core dynamics no explicit energy | 0.628 |

### Table B6. Low-energy decoding under energy-control feature sets.

| Feature set | Mean low-energy ROC-AUC |
|---|---:|
| Observation | 1 |
| Internal all | 0.957 |
| Internal dynamics | 0.952 |
| Explicit allostatic variables | 0.934 |
| internal dynamics no explicit energy | 0.927 |
| Core dynamics | 0.916 |
| core dynamics no explicit energy | 0.886 |
| Trace only | 0.798 |
| Policy head only | 0.682 |
| Observation without energy | 0.663 |

### Table B7. Functional-state decoding from internal dynamics.

| Functional target | Mean ROC-AUC |
|---|---:|
| Contact opportunity | 0.984 |
| Low energy | 0.949 |
| Threat context | 0.845 |
| Resource approach | 0.83 |
| Future contact within five steps | 0.629 |

### Table B8. Mean behavioral deltas by perturbation.

| Perturbation | Mean delta FSE | Mean delta no-resource failure | Mean delta reward | Rows aggregated |
|---|---:|---:|---:|---:|
| No membrane persistence | -0.447 | 0.43 | -22.357 | 30 |
| Temperature = 1.00 | -0.408 | 0.457 | -20.603 | 30 |

| Perturbation | Mean delta FSE | Mean delta no-resource failure | Mean delta reward | Rows aggregated |
|---|---|---|---|---|
| Contact sensor zeroed | -0.373 | 0.368 | -18.947 | 30 |
| Zero trace | -0.313 | 0.002 | -7.34 | 30 |
| No recurrent current | -0.246 | 0.222 | -11.445 | 30 |
| Zero trace + reset state | -0.24 | 0.241 | -10.633 | 30 |
| Reset state each step | -0.214 | 0.168 | -9.499 | 30 |
| No allostatic modulation | -0.038 | 0.039 | -2.124 | 30 |
| Zero contact trace | -0.028 | 0.002 | 0.012 | 30 |
| No oscillatory gate | -0.022 | 0.024 | -1.438 | 30 |
| Energy-neutral observation | -0.008 | -0.025 | 0.371 | 30 |

## Table B9. Fraction of perturbation rows with behavior and internal-prediction degradation.

| Perturbation | Fraction of model rows |
|---|---|
| Temperature = 1.00 | 0.6 |
| Zero trace + reset state | 0.4 |
| No allostatic modulation | 0.4 |
| Zero contact trace | 0.2 |
| No oscillatory gate | 0.2 |
| Energy-neutral observation | 0.2 |
| Contact sensor zeroed | 0.2 |

| Perturbation | Fraction of model rows |
|---|---|
| No recurrent current | 0.2 |
| No membrane persistence | 0.2 |
| Zero trace | 0 |
| Reset state each step | 0 |

## Table B10. Event target coverage.

| Target | Rows | Positive | Negative | Positive rate |
|---|---|---|---|---|
| Future contact within five steps | 825000 | 407094 | 417906 | 0.493 |
| Future successful consume within five steps | 825000 | 320552 | 504448 | 0.389 |
| Future threat hit within five steps | 825000 | 211345 | 613655 | 0.256 |
| Current low energy | 825000 | 86052 | 738948 | 0.104 |

## Table B11. Baseline internal-dynamics event prediction.

| Target | Mean baseline ROC-AUC |
|---|---|
| Current low energy | 0.914 |
| Future successful consume within five steps | 0.648 |
| Future contact within five steps | 0.602 |
| Future threat hit within five steps | 0.577 |

## Table B12. Perturbation sensitivity of future-event prediction.

| Perturbation | Mean future-event AUC delta |
|---|---|
| Zero trace | -0.091 |
| Zero trace + reset state | -0.076 |

| Perturbation | Mean future-event AUC delta |
| --- | --- |
| Temperature = 1.00 | -0.071 |
| Zero contact trace | -0.059 |
| No recurrent current | -0.052 |
| Contact sensor zeroed | -0.042 |
| Reset state each step | -0.029 |
| No allostatic modulation | -0.024 |
| Energy-neutral observation | -0.021 |
| No membrane persistence | -0.02 |
| No oscillatory gate | -0.012 |
| Baseline | 0 |

# Appendix C. Reproducibility and Artifact Manifest

Appendix C summarizes how the main manuscript maps onto the final frozen output artifacts. It is intended as a lightweight reproducibility guide rather than a complete dump of every intermediate log.

## C.1 Experiment-to-artifact mapping

| Paper component | Source output family | Final-output role |
| --- | --- | --- |
| Behavioral benchmark | Experiment 1 final behavioral outputs | Frozen held-out behavior, model/variant performance, robustness ranking |
| Internal prediction and compression | Experiment 2 internal-state outputs | Early FSE probes, permutation baselines, PCA compression |
| Feature-family and allostatic controls | Experiment 3 bridge-control outputs | Feature-family FSE probes, low-energy controls, functional-state probes |
| Causal perturbations | Experiment 4 causal perturbation outputs | Behavioral deltas, internal-prediction deltas, causal-bridge summaries |

| Paper component | Source output family | Final-output role |
|---|---|---|
| Future-event analysis | Experiment 5 seed-balanced event outputs | Event target coverage, baseline event probes, event perturbation deltas |
| Figure generation | Paper figure polish notebook | Final Figures 1-6 and Table 3 evidence matrix |

## C.2 Main manuscript figure/table manifest

| Artifact family | Contents | Location |
|---|---|---|
| Main figures | Figures 1–6 as PNG/PDF/SVG | outputs/figures/paper_main/ |
| Main tables | Table 3 and figure/table manifest | outputs/tables/paper_main/ |
| Appendix tables | Compact appendix CSV/Markdown tables | outputs/tables/paper_appendix/ |
| Appendix text | Combined appendix Markdown | outputs/appendices/paper_appendix/ |
| Full output bundle | Complete fold-level and artifact outputs | project output archive |

## C.3 Additional detailed manifests

The complete figure/table manifest and appendix source manifest are retained in the project output bundle. The main generated artifacts are the six paper figures, Table 3 evidence matrix, and compact appendix tables. Full paths, file sizes, fold-level outputs, complete perturbation rows, and event-sample summaries are not reproduced here in order to keep the manuscript readable.

## C.4 Code and data availability

Code, derived analysis tables, figure-generation scripts, and reproducibility manifests associated with this study are publicly available in the project repository at https://github.com/fehayes/predictive-allostatic-organization. The repository contains the materials needed to trace the reported manuscript results to the frozen analysis outputs and figure-generation workflow. Full lower-level outputs that are impractical to reproduce in the manuscript are retained according to the output-bundle policy described below. The main manuscript reports only frozen final analyses; pilot experiments were used for protocol development and are not used as final quantitative evidence.

## C.5 Output bundle policy

Full fold-level probe outputs, complete perturbation rows, raw event-sample summaries, and figure source files are retained in the project output bundle rather than reproduced in the PDF. The appendix includes compact tables sufficient to audit the main claims while keeping the manuscript readable.